\documentclass[]{spie}  %>>> use for US letter paper
\usepackage{amsmath,amsfonts,amssymb}
\usepackage{graphicx}
\usepackage[colorlinks=true, allcolors=blue]{hyperref}
\usepackage{subcaption}
\usepackage{bm}

\title{Protection of SVM Model with Secret Key \\
from Unauthorized Access}

\author[a]{Ryota Iijima}
\author[b]{MaungMaung AprilPyone}
\author[c]{Hitoshi Kiya}
\affil[a, b, c]{Tokyo Metropolitan University, 6-6 Asahigaoka, Hino, Tokyo, Japan, 191-0065}
\authorinfo{Further author information: \\R.Iijima: E-mail: iijima-ryota@ed.tmu.ac.jp, \\  MaungMaung AprilPyone: E-mail: april-pyone-maung-maung@ed.tmu.ac.jp, \\ Hitoshi Kiya: E-mail: kiya@tmu.ac.jp}

\begin{document} 
\maketitle

\begin{abstract}
In this paper, we propose a block-wise image transformation method with a secret key for support vector machine (SVM) models. Models trained by using transformed images offer a poor performance to unauthorized users without a key, while they can offer a high performance to authorized users with a key. The proposed method is demonstrated to be robust enough against unauthorized access even under the use of kernel functions in a facial recognition experiment. 
\end{abstract}


\section{INTRODUCTION}
The spread of machine learning has greatly contributed to solving complex tasks for many applications. Machine learning utilizes a large amount of data to extract representations of relevant features and rich computing resources, so a trained model should be regarded as a kind of intellectual property. \par
Accordingly, a model has to be protected from unauthorized access by an
illegal party who may obtain the model and use it for its own service \cite{maungmaung_kiya_2021, Fan2019RethinkingDN, 9454280}. In this paper, we propose a block-wise transformation with a secret key for protecting support vector machine (SVM) models. Users without a key cannot use the performance of the protected models. Conventional protection methods for deep neural network (DNN) models\cite{maungmaung_kiya_2021} are also demonstrated to be unavailable for protecting SVM models. Deep learning requires using is suitable for large and complex tasks, but it requires using both a large amount of data and rich computational resources. Therefore, models without deep learning such as SVM models become more important in applications that cannot meet such requirements. In addition to SVM, the proposed method can be applied to other machine leaning algorithms based on the Euclidean distance or the inner product between vectors.

\section{SVM}
SVM models have widely been used in data classification tasks. In SVM computeing, we input a feature vecotr $\bm{x}$ to the discriminant function as 
\begin{equation}
    f(\bm{x}) = \mathrm{sign}{\left ( \bm{\omega}^{T} \bm{x} + b \right )}
\end{equation}
with
\begin{equation}
    \mathrm{sign}{\left ( u \right )} = \left \{
    \begin{aligned}
    1 \; & (u > 1) \\
    -1 \; & (u \leq 0)
    \end{aligned}
    \right . , \nonumber
\end{equation}
where $\bm{\omega}$ is a weight parameter vector, $b$ is a bias, and $T$ indicates transpose. \par
The SVM has also a technique called "kernel trick". When the kernel trick is applied to eq.(1), the equation is given by 
\begin{equation}
    f(\bm{x}) = \mathrm{sign}{\left ( \bm{\omega}^{T} \phi \left ( \bm{x} \right ) + b  \right )}.
\end{equation}
The function $\phi(\bm{x}) : \mathbb{R}^d \rightarrow \mathcal{F}$ maps an input vector $\bm{x}$ on high dimensional feature space $\mathcal{F}$, where $d$ is the number of dimensions of features. The kernel function of two vectors $\bm{x}_i, \bm{x}_j$ is defined as
\begin{equation}
    K(\bm{x}_i, \bm{x}_j) = \left \langle \phi(\bm{x}_i), \phi(\bm{x}_j) \right \rangle, 
\end{equation}
where $\langle .,. \rangle$ is an inner product.
Typical kernel functions such as the radial basis function (RBF) kernel, linear one, and polynomial one are based on the distance or the inner products. For examples, the RBF kernel is based on the Euclidean distance and the polynomial kernel is based on the inner products, as
\begin{equation}
    K(\bm{x}_i,\bm{x}_j) = \exp{\left( -\gamma \left\| \bm{x}_i-\bm{x}_j \right\|^2 \right)}
\end{equation}

\begin{equation}
    K(\bm{x}_i,\bm{x}_j) = \left( 1+\langle \bm{x}_i,\bm{x}_j \rangle \right)^l
\end{equation}

\noindent
where $\bm{x}_i$ and $\bm{x}_j$ are input vectors, $\gamma$ is the hyperparameter for deciding the complexity of boundary determination, and $l$ is a parameter for deciding the degree of the polynomial. \par

\section{Proposed Method}

\subsection{overview}
Figure 1 shows the framework of the proposed model protection. Training data is transformed by using a block-wise transformation with a secret key $K$ \cite{aprilpyone2021block}. For testing, test data is transformed with key $K$ as well, and the transformed test data is applied to the model. In this framework, the trained models are expected not only to offer a degraded performance of models to unauthorized users without key $K$ but also to provide a high performance to authorized users with key $K$. To achieve this purpose, we propose applying a bock-wise transformation \cite{chuman2018encryption, TatsuyaCHUMAN20182017MUP0001}. The Block-wise transformation consists of four transformation steps (block permutation, pixel shuffling, bit flipping, z-score normalization) shown in Fig. 1(b). The detail of each step is presented as below. 

\begin{figure}
    \begin{tabular}{cc}
        \begin{minipage}[b]{0.5\linewidth}
            \centering
            \includegraphics[keepaspectratio, scale=0.25]{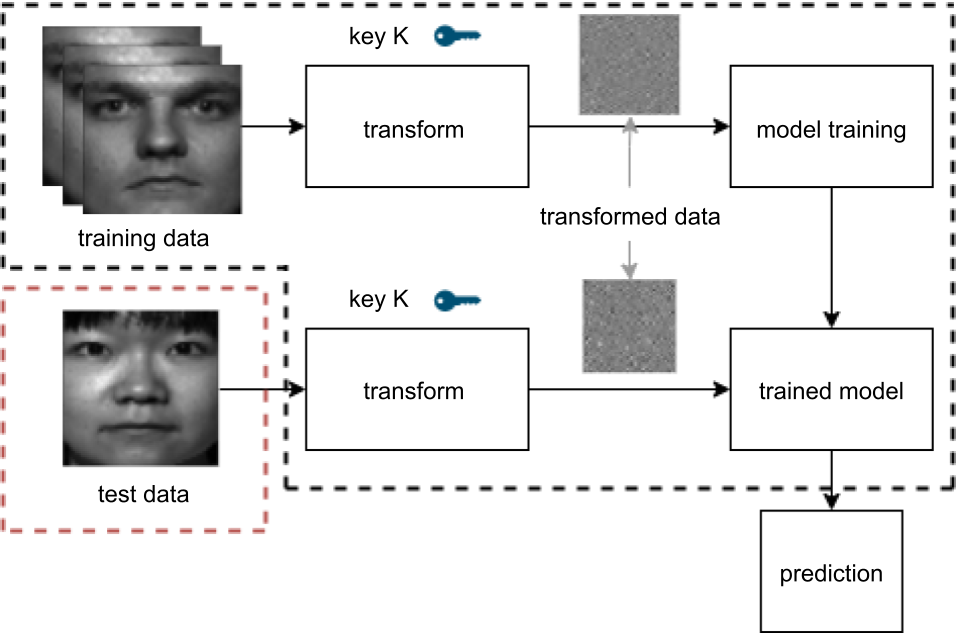}
            \subcaption{}
        \end{minipage}
        \begin{minipage}[b]{0.5\linewidth}
            \centering
            \includegraphics[keepaspectratio, scale=0.3]{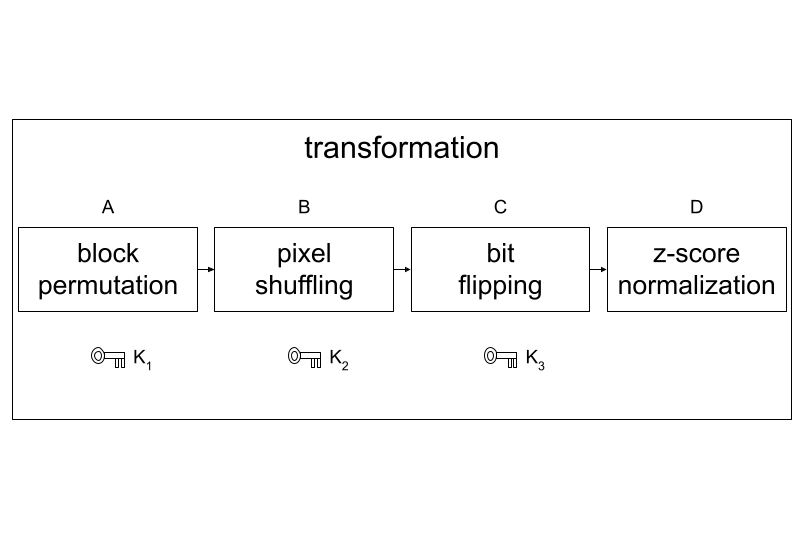}
            \subcaption{}
        \end{minipage}
    \end{tabular}
    \caption{Framework of our proposed method: (a)image classification with a secret key (b)transformation steps}
\end{figure}
% \begin{figure}
%     \centering
%     \includegraphics[keepaspectratio, scale=0.4]{figures/tifs-segmentation-copy2.png}
%     \vspace{3mm}
%     \caption{Process of block segmentation for transformation with a secret key}
%     \label{fig:my_label}
% \end{figure}

\subsection{Transformation with key}

The proposed transformation consists of four steps as follows (see Fig. 2).

\begin{figure}
    \begin{tabular}{ccccc}
        \begin{minipage}[b]{0.19\linewidth}
            \centering
            \includegraphics[keepaspectratio, scale=0.32]{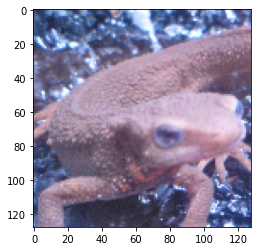}
            \subcaption{}
        \end{minipage}
        \begin{minipage}[b]{0.19\linewidth}
            \centering
            \includegraphics[keepaspectratio, scale=0.32]{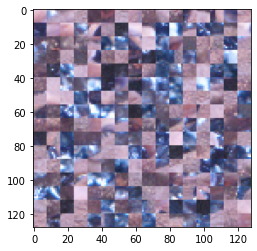}
            \subcaption{}
        \end{minipage}
        \begin{minipage}[b]{0.19\linewidth}
            \centering
            \includegraphics[keepaspectratio, scale=0.32]{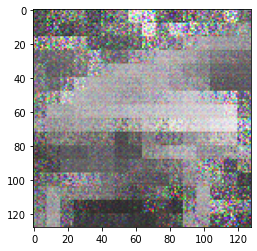}
            \subcaption{}
        \end{minipage}
        \begin{minipage}[b]{0.19\linewidth}
            \centering
            \includegraphics[keepaspectratio, scale=0.32]{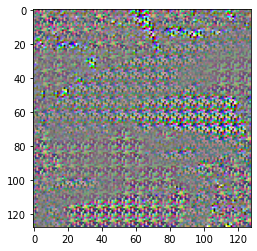}
            \subcaption{}
        \end{minipage}
        \begin{minipage}[b]{0.19\linewidth}
            \centering
            \includegraphics[keepaspectratio, scale=0.32]{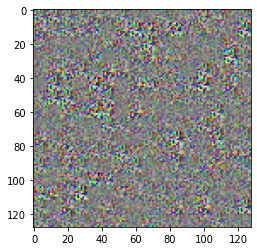}
            \subcaption{}
        \end{minipage}
    \end{tabular}
    \caption{Examples of transformed image: (a)original (b)block permutation (c)pixel shuffling (d)bit flipping (e)block permutation, pixel shuffling, and bit flipping}
    \vspace{3mm}
\end{figure}

\begin{figure}
    \centering
    \includegraphics[keepaspectratio, scale=0.4]{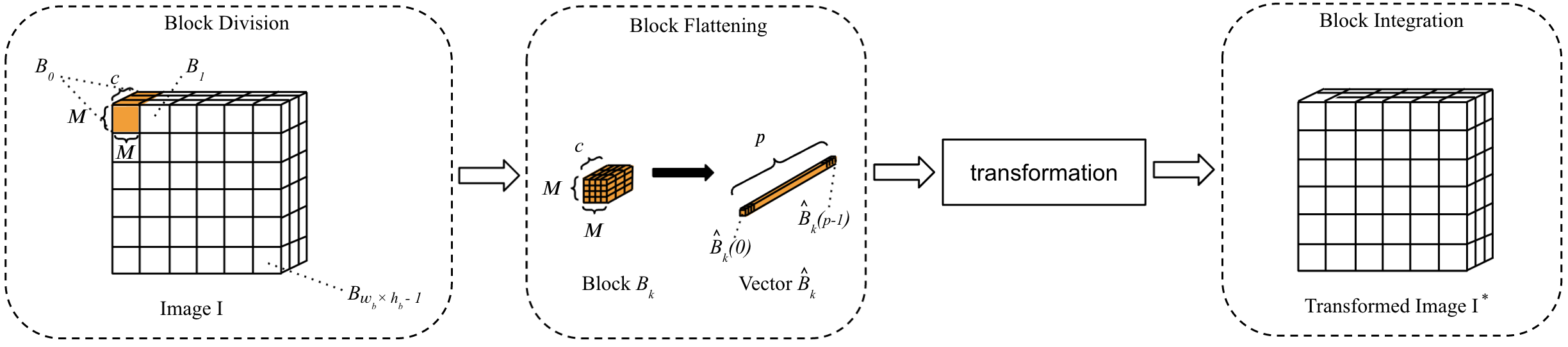}
    \vspace{3mm}
    \caption{Process of block segmentation for transformation with a secret key}
    \label{fig:my_label}
\end{figure}

\subsubsection*{A. block permutation}
\begin{enumerate}
    \item An input image $I$ with a size of $W \times H \times C$ is divided into blocks $B_{k}, k \in \{ 0,1,...,w_b \times h_b - 1 \} $ with a size of $M \times M \times C$, and each block is assigned an index such that
    \begin{equation}
        B = \left ( B_0,B_1,...B_{k},...,B_{w_b \times h_b - 1} \right ),
    \end{equation}
    where $H$, $W$, and $C$ are the height, width, and the number of channels of an image, and $M$ is the block size of transformation (see Fig. 3). Therefore, $h_b=H/M$ $\times$ $w_b=W/M$ blocks are defined. For simplicity, let $H$ and $W$ be integers multiples of $M$. 
    
    \item Generate a random permutation vector $v = (v_0, v_1, ..., v_k, ..., v_{k'}, ..., v_{h_b \times w_b -1})$ that consists of randomly permuted integers form $0$ to $h_b \times w_n - 1$ by using a key $K_1$, where $k, k' \in \{0, ..., h_b \times w_b - 1 \}$, and $v_k \neq v_{k'}$ if $k \neq k'$.
    \item Blocks are shuffled by the vector $v$, i.e.,
    \begin{equation}
        B(k)=B(v_k).
    \end{equation}
    % where $B(k)$ is the k-th block in $B$ (see eq.(6)).
\end{enumerate}

\subsubsection*{B. pixel shuffling}
\begin{enumerate}
    \item Each block $B_{k}$ is flattened to  a vector $\hat{B}_{k} = (\hat{B}_{k}(0),\hat{B}_{k}(1),...,\hat{B}_{k}(l),...,\hat{B}_{j\,j}(p-1))$, where $\hat{B}_{k}(l)$ is the $l$-th pixel value in $\hat{B}_{k}$, $l \in \{0, ..., p-1 \}$, and $p = M \times M \times C$ is the number of pixels in a block (see Fig. 3).
    \item Generate a random permutation vector $v = (v_0, v_1, ..., v_l, ..., v_{l'}, ..., v_{p-1})$ that consists of randomly permuted integers form $0$ to $p - 1$ by using a key $K_2$, where $l, l' \in \{0, ..., p-1 \}$ and $v_l \neq v_{l'}$ if $l \neq l'$.
    \item Pixel values in a vector $\hat{B}_{k}$ are shuffled by the vector $v$, i.e.,
    \begin{equation}
        \hat{B}_{k}(l)=\hat{B}_{k}(v_l). \nonumber
    \end{equation}
    
\end{enumerate}

\subsubsection*{C. bit flipping}
\begin{enumerate}
    \item A binary vector $r = \left ( r_0, r_1, ..., r_l, ... r_{p-1} \right ), r_l \in \left\{ 0, 1 \right\}$ is generated by a key $K_3$, where $r$ consists of 50\% of "1" and 50\% of "0".
    \item Pixel values in $\hat{B}_k$ are transformed by negative-positive (NP) transformation on the basis of $r$ as
    \begin{gather}
        \hat{B}_{k}(l) = \left\{ 
        \begin{aligned}
        & \hat{B}_{k}(l) && (r_l = 0) \\
        & \hat{B}_{k}(l) \oplus (2^L-1) && (r_l = 1)
        \end{aligned}
        \right. \nonumber,
    \end{gather}
    where $L$ is the number of bits used for $\hat{B}_{k}(l)$, and $L=8$ is used in this paper. 
\end{enumerate}

\subsubsection*{D. z-score normalization}
Let us transform an image $I_i, i \in \{1,2,...,N\}$ with $W \times H$ pixels into
a vector $\hat{I}_i = (\hat{I}_i(0),\hat{I}_i(1),...,\hat{I}_i(t),...,\hat{I}_i(H \times W \times C - 1)), t \in \{ 0,1,...,H \times W \times C - 1 \}$, where $N$ is the number of images and $\hat{I}_i(t)$ is the $t$-th pixel value in $\hat{I}_i$. z-score normalization is applied to $I_i(t)$ as in 
\begin{equation}
    z_i(t) = \frac{\hat{I}_i(t) - \sigma_t}{\mu_t}
\end{equation}
where $\sigma_t$ is the mean value of the data given by 
\begin{equation}
    \sigma_t = \frac{1}{N} \sum^N_{i=1}\hat{I}_i(t),
\end{equation}
and $\mu_t$ is the standard deviation given by
\begin{equation}
    \mu_t = \sqrt{\frac{\sum^N_{i=1}(\hat{I}_i(t) - \sigma_t)^2}{N}}.
\end{equation}

After z-score normalization, all blocks are integrated to form a transformed image $I^*$ (see Fig. 3).

\subsection{Properties of transformed images}
To confirm the properties of transformed images, we focus on the Euclidean distance and the inner product between two vectors.

\subsubsection*{A. block permutation and pixel shuffling}
Block permutation and pixel shuffling are expressed by using an orthogonal matrix $Q$ as
\begin{equation}
    \hat{I}^*_i = Q \hat{I}_i.
\end{equation}
Therefore, these transformations meet the properties in eq.(11) and eq.(12). 
\\
Property 1 : Conservation of Euclidean distances
\begin{equation}
    \left\| \hat{I}^*_i - \hat{I}^*_j \right\|^2 = \left\| \hat{I}_i - \hat{I}_j \right\|^2
\end{equation}
Property 2 : Conservation of inner products
\begin{equation}
    \langle \hat{I}^*_i, \hat{I}^*_j \rangle = \langle \hat{I}_i, \hat{I}_j \rangle,
\end{equation}
where $I^*_i$ and $I^*_j$ are images transformed from $I_i$ and $I_j$, and $\hat{I}^*_i$ and $\hat{I}^*_j$ are the vector expression of $I^*_i$ and $I^*_j$, respectively. Accordingly, the use of block permutation and pixel shuffling do not affect the performance of SVM models with the RBF kernel (eq.(4)) and the polynomial kernel (eq.(5)).

\subsubsection*{B. bit flipping}
Images transformed with bit flipping meet property 1, but they do not satisfy property 2 as bellow.
\begin{align}
    \left\| \hat{I}^*_i(t) - \hat{I}^*_j(t) \right\|^2 &= \left\| (255 - \hat{I}_i(t)) - (255 - \hat{I}_j(t) \right\|^2 \nonumber \\
    &= \left\| \hat{I}_i(t) - \hat{I}_j(t) \right\|^2,
\end{align}
\begin{align}
    \langle \hat{I}^*_i(t), \hat{I}^*_j(t) \rangle &= \langle (255 - \hat{I}_i(t)), (255 - \hat{I}_j(t) \rangle \nonumber \\
    &\neq \langle \hat{I}_i(t), \hat{I}_j(t) \rangle ,
\end{align}
where $\hat{I}^*_{i}(t)$ and $\hat{I}^*_j(t)$ are the $t$-th pixel values in $\hat{I}^*_{i}$ and $\hat{I}^*_j$. Therefore, bit flipping is expected to maintain only the performance of SVM models with the RBF kernel. 

\subsubsection*{C. bit flipping with z-score normalization}
To solve the problem that the use of bit flipping cannot meet property 2, bit flipping with z-score normalization is applied to $\hat{I}_i$ as bellow.
\begin{align}
    z^*_i(t) &= \frac{(255 - \hat{I}_i(t)) - (255 - \sigma_t)}{\mu^*_t} \nonumber \\
    &= - \frac{\hat{I}_i(t) - \sigma_t }{\mu_t} = - z_i(t),
\end{align}
where 
\begin{align}
    \mu^*_t &= \sqrt{\frac{\sum^N_{i=0}((255 - \hat{I}_i(t)) - (255 - \sigma_t))^2}{N}} \nonumber \\
    &= \sqrt{\frac{\sum^N_{i=0}(\sigma_t - \hat{I}_i(t))^2}{N}} = \mu_t.
\end{align}
Therefore, under the use of z-score normalization, SVM can meet the both properties 1 and 2, even when the polynomial kernel is used \cite{kawamura2020privacy, Nakamura2016UnitaryTT} .
% \begin{figure}
%     \begin{tabular}{ccccc}
%         \begin{minipage}[b]{0.19\linewidth}
%             \centering
%             \includegraphics[keepaspectratio, scale=0.32]{figures/imori_original.png}
%             \subcaption{}
%         \end{minipage}
%         \begin{minipage}[b]{0.19\linewidth}
%             \centering
%             \includegraphics[keepaspectratio, scale=0.32]{figures/imori_block_scrambling.png}
%             \subcaption{}
%         \end{minipage}
%         \begin{minipage}[b]{0.19\linewidth}
%             \centering
%             \includegraphics[keepaspectratio, scale=0.32]{figures/imori_pixel_shuffling.png}
%             \subcaption{}
%         \end{minipage}
%         \begin{minipage}[b]{0.19\linewidth}
%             \centering
%             \includegraphics[keepaspectratio, scale=0.32]{figures/imori_bit_flipping.png}
%             \subcaption{}
%         \end{minipage}
%         \begin{minipage}[b]{0.19\linewidth}
%             \centering
%             \includegraphics[keepaspectratio, scale=0.32]{figures/imori_etc.png}
%             \subcaption{}
%         \end{minipage}
%     \end{tabular}
%     \caption{Examples of transformed image: (a)original (b)block permutation (c)pixel shuffling (d)bit flipping (e)block permutation, pixel shuffling, and bit flipping}
% \end{figure}

\section{Experiment}
\subsection{Experimental Setup}
We used the Extended Yale Face Data Base B \cite{georghiades2001few}, which consisits of $38 \times 64$ front face images with a size of $192 \times 168$ pixels for 38 persons. We resized all images to images with a size of $50 \times 50$ pixels, and divided the dataset into $38 \times 51$ and $38 \times 13$ images for training and testing. The RBF kernel and the polynomial kernel were used as a kernel function for SVM models. The hyper parameters of SVM models with the RBF kernel were $C = 512$ and $\gamma = 0.0001$, and the hyper parameters of SVM models with the polynomial kernel were $C=512$, $\gamma = 0.001$, and $degree = 2$. The block-wise transformation with a secret key $K$ was carried out with four steps as in Fig. 1, where $M=2$ and $M=5$ were selected as a block size.

\subsection{Effects of Proposed Method}
Tables 1 and 2 show experiment results of our proposed method. "with key" in the tables means that the test data was transformed with key $K$, and "without key" means that the test data was not transformed. 
From the tables, the proposed models with key were demonstrated to have the same accuracy of corresponding baselines, which was calculated by using original image. In contrast, unauthorized users without key cannot use the performance of the models. Accordingly, the proposed method enables us to achieve access control for SVM models.
\begin{table}[h]
    \begin{minipage}[t]{.45\textwidth}
        \caption{Experiment result for SVM with RBF kernel}
        \begin{center}
            \begin{tabular}{c|cc}
               \hline
               transform & with key & without key \\
               \hline \hline
               proposed(M=2) & 0.9757 & 0.0283\\
               proposed(M=5) & 0.9757 & 0.0303 \\
               \hline
               \hline
               baseline & \multicolumn{2}{c}{0.9757} \\
               \hline
            \end{tabular}
        \end{center}
        \label{left label}
    \end{minipage}
    \hfill
    \begin{minipage}[t]{.45\textwidth}
        \caption{Experiment result for SVM with polynomial kernel}
        \begin{center}
            \begin{tabular}{c|cc}
               \hline
               transform & with key & without key \\
               \hline \hline
               proposed(M=2) & 0.9473 & 0.0263\\
               proposed(M=5) & 0.9473 & 0.0263 \\
               \hline
               \hline
               baseline & \multicolumn{2}{c}{0.9473} \\
               \hline
            \end{tabular}
        \end{center}
        \label{right label}
    \end{minipage}
\end{table}

\section{Conclusion}
In this paper, we proposed a block-wise transformation method with a secret key to achieve the access control of SVM models from unauthorized users. In the experiment, the models trained by images transformed with the proposed method offered a same performance of baseline to authorized users. In contrast, the models without key provide a degraded performance to unauthorized users. In addition, this method is also available for other machine learning algorithms such as random forest and k-means clustering, which are based on Euclidean distance and inner product.

% \subsection{key estimation}
% Key space is depend on the block size of the transformation, the number of channels in an image, and the size of an image. The key spaces of block permutation, pixel shuffling, and bit flipping are given by $(H_b \times W_b)!$, $p!$, and $\frac{p!}{(p/2)!(p/2)!}$, respectively.

% \acknowledgments % equivalent to \section*{ACKNOWLEDGMENTS}       
 
% This unnumbered section is used to identify those who have aided the authors in understanding or accomplishing the work presented and to acknowledge sources of funding.  

% References
% \bibliography{report} % bibliography data in report.bib
\bibliographystyle{spiebib} % makes bibtex use spiebib.bst
\bibliography{main}

\end{document}